\documentclass[letterpaper, 10 pt, journal, twoside]{IEEEtran}
\usepackage{amsmath,amsfonts}
\usepackage{array}
\usepackage[caption=false,font=normalsize,labelfont=sf,textfont=sf]{subfig}
\usepackage{textcomp}
\usepackage{stfloats}
\usepackage{url}
\usepackage{verbatim}
\usepackage{graphicx}
\usepackage{cite}

\usepackage{bm}
\usepackage{wrapfig}
\usepackage[ruled,vlined]{algorithm2e}
\usepackage[noend]{algpseudocode}
\usepackage{tabularx}
\usepackage{multirow, makecell}
\usepackage{amssymb}
\usepackage[table]{xcolor}
\usepackage{cancel}

\newcommand{\RNum}[1]{\uppercase\expandafter{\romannumeral #1\relax}}
\newcommand{\rnum}[1]{\expandafter{\romannumeral #1\relax}}

\newcommand{\rd}{{\mathrm d}}

\newcommand{\rT}{{\mathrm{T}}}

\newcommand{\va}{{\bf a}}

\newcommand{\vx}{{\bf x}}

\newcommand{\vg}{{\bf g}}

\newcommand{\vu}{{\bf u}}

\newcommand{\vv}{{\bf  v}}

\newcommand{\vF}{{\bf F}}

\newcommand{\E}{\mathbb{E}}

\newcommand{\Qb}{\mathbb{Q}}
\newcommand{\Pb}{\mathbb{P}}

\newcommand{\Rb}{\mathbb{R}}

\newcommand{\ExP}[2]{\E_{{#1}}{\left[#2\right]}}

\newcommand{\pluseq}{\mathrel{+}=}

\begin{document}

\title{Smooth Model Predictive Path Integral Control without Smoothing}

\author{Taekyung Kim, Gyuhyun Park, Kiho Kwak, Jihwan Bae, and Wonsuk Lee
\thanks{Manuscript received: February 24, 2022; Revised: June 4, 2022; Accepted: July 14, 2022. This paper was recommended for publication by Editor H. Kurniawati upon evaluation of the Associate Editor and Reviewers' comments. \textit{(Corresponding author: Wonsuk Lee.)}}%
\thanks{Taekyung Kim, Kiho Kwak, Jihwan Bae, and Wonsuk Lee are with the Ground Technology Research Institute, Agency for Defense Development, Daejeon 34186, Republic of Korea (e-mail: ktk1501@add.re.kr; khkwak@add.re.kr; jihan1008@add.re.kr; wsblues@add.re.kr).}%
\thanks{Gyuhyun Park is with the Department of Mechanical Engineering, Seoul National University, Seoul 08826, Republic of Korea and also with the Ground Technology Research Institute, Agency for Defense Development, Daejeon 34186, Republic of Korea (e-mail: khpark@add.re.kr).}%
\thanks{Digital Object Identifier (DOI): 10.1109/LRA.2022.3192800}
}
\markboth{IEEE Robotics and Automation Letters. Preprint Version. Accepted July, 2022}%
{Kim \MakeLowercase{\textit{et al.}}: Smooth Model Predictive Path Integral Control without Smoothing}


\maketitle

\begin{abstract}
We present a sampling-based control approach that can generate smooth actions for general nonlinear systems without external smoothing algorithms. Model Predictive Path Integral (MPPI) control has been utilized in numerous robotic applications due to its appealing characteristics to solve non-convex optimization problems. However, the stochastic nature of sampling-based methods can cause significant chattering in the resulting commands. Chattering becomes more prominent in cases where the environment changes rapidly, possibly even causing the MPPI to diverge. To address this issue, we propose a method that seamlessly combines MPPI with an input-lifting strategy. In addition, we introduce a new action cost to smooth control sequence during trajectory rollouts while preserving the information theoretic interpretation of MPPI, which was derived from non-affine dynamics. We validate our method in two nonlinear control tasks with neural network dynamics: a pendulum swing-up task and a challenging autonomous driving task. The experimental results demonstrate that our method outperforms the MPPI baselines with additionally applied smoothing algorithms.
\end{abstract}


\section{Introduction}
\IEEEPARstart{C}{ontrolling} an autonomous vehicle in complex environments is a challenging problem. When the vehicle drives on a structured road, it can be modeled as a linear or a kinematic system that is easy to solve. However, the nature of the real world dictates non-linearity and this characteristic is highlighted more at high vehicle speeds and low road surface friction levels. The majority of autonomous driving research has been focused on normal driving conditions, while areas of aggressive maneuvering in highly nonlinear environments have not been fully addressed by prior work.

\begin{figure}[t]
\centering
\includegraphics[width=0.99\linewidth]{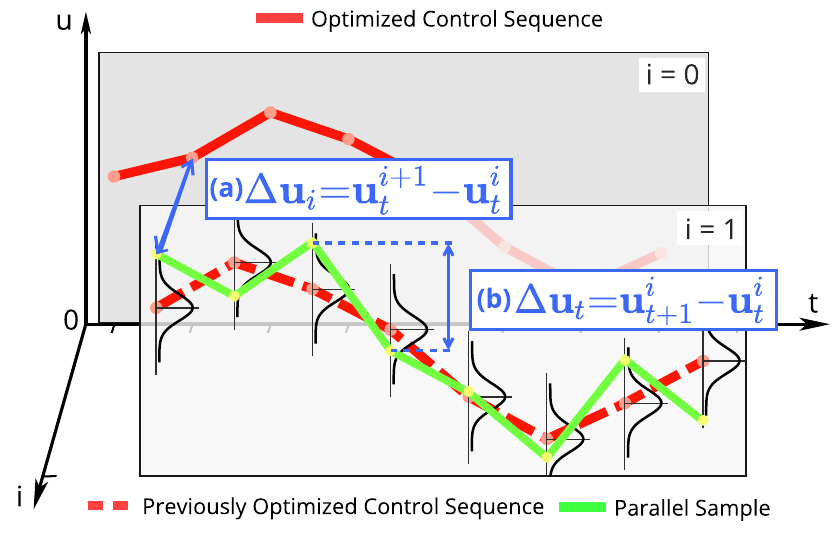}
\caption{A simplified representation of the MPPI algorithm during each optimization iteration. For clarity, we only visualize one sampled trajectory (in green). (a) Amount of changes between previously computed control sequence and the next control sequence (along the ``i-axis"). (b) Amount of changes in control values during MPPI rollouts (along the ``t-axis"), which are hard to be minimized by the MPPI baseline.}
\label{fig:changes}
\end{figure}

Gradient-based Model Predictive Control (MPC) methods have been introduced as powerful solutions for solving the problems of nonlinear control systems \cite{li2004iterative, tassa2014control}. Despite the fact that there have been successful real-world applications with these methods \cite{rosolia2020learning, kabzan2020amz}, they still possess the limitation that the cost function has to be differentiable. Sampling-based Model Predictive Path Integral (MPPI) control has been proposed to optimize both convex and non-convex objectives \cite{williams2017information} and has benefitted greatly from recent advances in Graphics Processing Units (GPUs), as a large number of samples can be computed in parallel to achieve better real-time performance. 

While MPPI presents general performances in various situations, it may encounter chattering when implemented on an actual platform, which is a common characteristic of sampling-based methods \cite{williams2018information}. Rapid changes in action commands are widely known to burden the actuators and cause the system to become unstable. Meanwhile, the key property of MPPI for achieving high-level control performance is to draw samples upon the previously optimized control sequence. However, this property can instead hinder the entire sequence from converging quickly to the optimal distribution when the environment changes too rapidly \cite{yinTrajectoryDistributionControl2022}. Under such circumstances, in particular, chattering becomes more prominent.

Therefore, a smoothing algorithm is often employed to smooth the control sequence in practice. However, because the optimization procedure and this external smoothing mechanism are completely decoupled, the smoothed sequence may unintentionally lose its optimality. In addition, it may present a significant problem if the algorithm behaves incorrectly, even if it is tuned to perform well in most situations. In the worst case, the resulting sequence may violate the physical constraints or diverge.

In this paper, we propose the Smooth Model Predictive Path Integral (SMPPI) control method that combines MPPI with an input-lifting strategy to generate smooth actions without any additional smoothing algorithms. We include derivative actions as control inputs, in which this technique is commonly used to smooth jerky commands in control studies \cite{vazquez2020optimization, chisari2021learning}. To the best of our knowledge, we seamlessly apply this idea in the MPPI framework for the first time while preserving the information theoretic interpretation of optimal control. Additionally, our method offers a new smoothing mechanism which is ineligible in the MPPI baseline. The smoothing effects of our approach are twofold. First, it smooths the gap between the control sequences of previous and current iterations. We refer to this as smoothing along the ``\textit{i-axis}." Second, it smooths the optimized control sequence during trajectory rollouts. We refer to this as smoothing along the ``\textit{t-axis}." The amount of changes in the control values of each axis that can be reduced by our method are depicted in Fig.~\ref{fig:changes}. Unlike MPPI, where efforts to realize smoothing along both axes would obstruct its ability to adapt to rapidly changing environments, SMPPI can achieve both smoothness and agility. We conduct experiments to compare our method with different smoothing methods in a classical control benchmark. We also demonstrate our idea in a challenging autonomous driving task. \footnote{Our video can be found at: \url{https://youtu.be/fyngK8PCoyM}}

\section{Background \label{section:mppi}}
This section provides a brief overview of the MPPI algorithm. The benefit of the sampling-based method is that it does not require a gradient of the objective function and the cost function \cite{kobilarov2012cross}. Recent works including MPPI have integrated a sampling-based approach with MPC formulations to solve non-convex optimization problems \cite{williams2017information, williams2018information}. They combine information theoretic control and stochastic optimal control using free energy and the KL divergence to derive an optimal control distribution with importance sampling \cite{theodorou2012relative}.

Consider a discrete time dynamic system with a state $\vx_t \in \Rb^n$ and a control input $\vv_t \in \Rb^m$. We applied the general noise assumption that holds that we could not directly control the system over $\vv_t$, but could over the mean $\vu_t$ of the density function of noise $\bm{\epsilon}_t$:
\begin{equation}
\vv_t \sim \mathcal{N}(\vu_t, \bm{\Sigma}). \nonumber
\end{equation}
Noise is simply defined as $\vv_t = \vu_t + \bm{\epsilon}_t$. Given a sequence of inputs $V = \{\vv_0, \vv_1, \dots \vv_{T-1}\}$ and mean input variables $U = \{\vu_0, \vu_1, \dots \vu_{T-1}\}$ along a finite time horizon $t \in \{0, 1, \dots T-1 \}$, we can define the probability density function $q(V)$ as:
\begin{equation}
q(V) = \prod_{t=0}^{T-1} Z^{-1}\exp\left( -\frac{1}{2}(\vv_t - \vu_t)^\rT \bm{\Sigma}^{-1} (\vv_t - \vu_t)\right) ,   
\label{Equation:controlled}
\end{equation}
where $Z = \left( (2\pi)^m |\bm{\Sigma}| \right)^{\frac{1}{2}}$. Similarly, we can also define the uncontrolled density function $p(V)$ where $U$ is usually 0:
\begin{equation}
p(V) = \prod_{t=0}^{T-1} Z^{-1}\exp\left( -\frac{1}{2}\vv_t^\rT \bm{\Sigma}^{-1} \vv_t\right) .   
\label{Equation:uncontrolled}
\end{equation}
These two density functions correspond to the distributions of $\Qb$ and $\Pb$, respectively. Here, we define an optimal density function using the free energy of the system \cite{theodorou2012relative}, which corresponds to the optimal distribution $\Qb^*$:
\begin{equation}
q^*(V) = \frac{1}{\eta}\exp\left(-\frac{1}{\lambda}S(V) \right) p(V) , \label{Equation:optimal}
\end{equation} 
where $\eta$ denotes the normalizing constant and $S(V)$ denotes the state-dependent cost. The input sequence in state cost is iteratively transformed into state values $\vx$ through the non-affine system dynamics $\vF$ \cite{williams2018information}:
\begin{align}
    S(V;\vF) &= \phi(\vx_T) + \sum_{t=0}^{T-1} c(\vx_t) , \label{Equation:state_cost}\\
    \vx_{t+1} &= \vF(\vx_t, \vv_t) . \nonumber
\end{align}
As proposed in \cite{williams2018information}, we can now derive the optimal control input by minimizing the KL divergence between $\Qb$ and $\Qb^*$:
\begin{equation}
\ExP{\Qb^*}{\vv_t} = \int q^*(V) \vv_t \rd V .
\label{Equation:optimal_control}
\end{equation}
Here, the best solution of (\ref{Equation:optimal_control}) is to draw samples directly over $\Qb^*$, but this is not possible. Importance sampling is therefore employed to compute the integral over the known distribution $\Qb$:
\begin{align}
\ExP{\Qb^*}{\vv_t} &= \int {w(V)} {q(V)} \vv_t \rd V \nonumber \\
&= \ExP{\Qb_{U,\bm{\Sigma}}}{w(V) \vv_t} ,
\end{align}
where the importance weighting term is:
\begin{align}
&w(V) = \left( \frac{q^*(V)}{p(V)} \right) \left( \frac{p(V)}{q(V)} \right) \nonumber \\ 
 &= \frac{1}{\eta}\exp\left( -\frac{1}{\lambda}\left(S(V) + \frac{1}{2}\lambda\sum_{t=0}^{T-1} \vu_t^\rT \bm{\Sigma}^{-1} (\vu_t + 2\bm{\epsilon}_t)  \right) \right) . 
\end{align}
Let $\mathcal{E}$ be a noise sequence $\{\bm{\epsilon}_0, \bm{\epsilon}_1, \dots {\bm{\epsilon}_{T-1}}\}$ and $K$ be the number of trajectory samples. Finally, we have the iterative optimal control update law to compute the weights when the sampled trajectory cost $\{C(V^0), C(V^1), \dots C(V^{K-1})\}$ is given:
\begin{align}
\vu_t^{i+1} &= \vu_t^{i} + \sum_{k=0}^{K-1} w(\mathcal{E}^k) \bm{\epsilon}_t^k  \label{Equation:no_filter} \\
w(\mathcal{E}^k) &= \frac{1}{\eta}\exp\left( -\frac{1}{\lambda}\left(C(V^k) - \beta \right) \right),
\label{Equation:update_law}
\end{align}
where we subtract the minimum state cost $\beta$ to ensure that at least one sample has a numerically non-zero importance sampling weight.
The trajectory cost in (\ref{Equation:update_law}) is then simplified as follows:
\begin{equation}
C(V^k) = S(V^k) + \lambda\sum_{t=0}^{T-1} \vu_t^\rT \bm{\Sigma}^{-1} \bm{\epsilon}_t^k.
\label{Equation:traj_cost}
\end{equation}

\section{Smooth Model Predictive Path Integral Control}
\subsection{Limitations of External Smoothing Algorithms}

The sampling-based approach of the MPPI algorithm makes possible the consideration of involving non-convex objectives in optimization problems. However, the stochastic nature of this method can cause significant chattering in the resulting commands. Due to the unique property by which perturbed trajectories are sampled around the previously optimized mean control sequence $U$, MPPI cannot respond effectively when the optimal control distribution greatly deviates from the previous iteration. Undesirable chattering effects stand out more in the scenarios mentioned above.

A common approach to mitigate such a problem is to smooth the resulting control sequence using, for instance, sliding window smoothing methods or filtering based methods \cite{sarkka2008unscented, ruiz2017particle}. It was suggested that a Savitzky-Golay filter (SGF) \cite{savitzky1964smoothing} would be an effective solution for this control method \cite{williams2018information}. It smooths the subsets of adjacent data by fitting the local polynomial approximations in a convolutional manner.

However, this type of approach can have negative effects, as external manipulations of control values are not considered in the optimization process. Consider a general state-action system dynamics with clamping function $\vg_{\vu}$ for handling control constraints: 
\begin{equation} 
    \vx_{t+1} = \vF\left(\vx_t, \vg_{\vu}(\vv_t)\right).
    \label{Equation:dynamics}
\end{equation}
Because the perturbed control $\vv_t$ is bounded by the physical limits of the actuators, the sampled noise should be bounded as follows:
\begin{equation}
    \bar{\bm{\epsilon}}_t = \vg_{\vu}(\vv_t) - \vu_t.
\end{equation}
Two possible ways exist to apply the SGF in the control sequence. 

\subsubsection{Smoothing weighted noise sequence}
First, we can apply smoothing to the weighted noise sequence:
\begin{equation}
    \vu_t^{i+1} = \vu_t^{i} + \operatorname{SGF}\left(\sum_{k=0}^{K-1} w(\bar{\mathcal{E}}^k) \bar{\bm{\epsilon}}_t^{k} \right) . \label{Equation:filter_on_noise}
\end{equation}
This makes data points smoother while compensating for the noise, which changes considerably along the time horizon. During this procedure, the control variable $\vu$ may violate the constraint conditions given that the bounded noises are manipulated by the filter. On the other hand, the nature of this process makes it vulnerable to phase distortion \cite{adhikari2020physiological}. In addition to violating the given constraints, the control sequence will diverge if the noises are overlapped repeatedly and amplified over time. We describe this phenomenon further in Section~\ref{subsection:result}.

\subsubsection{Smoothing control sequence}
Second, we can apply the SGF after the control sequence is updated using the weighted noise:
\begin{equation}
    \vu_t^{i+1} = \operatorname{SGF}\left(\vu_t^{i} + \sum_{k=0}^{K-1} w(\bar{\mathcal{E}}^k) \bar{\bm{\epsilon}}_t^{k} \right) . \label{Equation:filter_on_u}
\end{equation}
Although this method guarantees that the control variables are meeting the bounded condition, it can cause a delay in the system response. It is common to provide history values as they are required in the convolution-based smoothing method. The early values in the control sequence are then affected by the history, which means that polynomial approximation hinders the controlled distribution from rapidly shifting to an optimal distribution \cite{wang2014fractional}. Note that the first control action is the key value in the MPC problem.

\subsection{Decoupling Control Space and Action Space}

In the original MPPI, the running control cost in (\ref{Equation:traj_cost}) only plays a role in reducing the distance between control values of the current and previous iterations \cite{nakatani2019swing}. Chattering innately occurs because the control trajectory is sampled randomly, and the variance along the ``t-axis" in the new control trajectory is not taken into account during optimization. To this end, it is possible to impose an additional cost in terms of the variation of the input, i.e., chattering on the ``t-axis." \cite{williams2017model} suggested an additional running cost for variation of the inputs, but it cannot be applied generally, as it assumes that the dynamics are affine in control under special conditions. We intend to expand the MPPI baseline \cite{williams2018information}, which is likely to be adopted more extensively, but arbitrary modifications for the additional cost associated with input sequences violate the theoretical derivation of MPPI as the control cost is determined inherently through the importance sampling scheme.

In order to handle such an issue while preserving the MPPI framework, therefore, we suggest the lifting of the control variables as derivative actions. This strategy can mitigate the chattering problem in terms of two points of view.

\subsubsection{Smoothing along the ``i-axis"}

Henceforth, let us decouple the control space and action space by defining action sequence $A = \{\va_0, \va_1, \dots \va_{T-1}\}$. Noisy sampling is performed on a higher order control space $U$. Then, the resulting control sequence is integrated to become a smoother action sequence. The control distribution is hereby distinguished from the MPPI baseline. The variance of the injected noise is adjusted according to the physical limit of the input rate of change. This in turn reduces the variation along the ``i-axis" and consequently mitigates damage to the actuator.

Note that one can attempt to smooth resulting actions in the MPPI framework by reducing the variance in the same manner as described above. In this setting, however, MPPI would be unable to respond to rapidly changing environments. On the other hand, increasing the variance to cover a wider action range would not only be physically implausible but would also eventually result in chattering. The method proposed here is free from this trade-off, as its new search space $U$ can be handled adaptively depending on the environment.

\subsubsection{Smoothing along the ``t-axis"}
In contrast to MPPI, our control distribution corresponds to derivative action, allowing action variables to be treated as augmented state elements and hence making them independent of the control cost. Variation of action can now be easily included in the state cost such that chattering with regard to the ``t-axis" can be minimized. Here, we introduce an extra action cost $\Omega$ to smooth the action sequence along the ``t-axis" by minimizing the variance of $A$ without violating the information theoretic interpretation of MPPI:
\begin{equation} \label{eq:actionseq}
    \Omega(A) = \sum_{t=1}^{T-1} (\va_t - \va_{t-1})^{\rT} {\bm{\omega}} \, (\va_t - \va_{t-1}) ,
\end{equation}
where $\bm{\omega}$ is the weighting parameter in the form of a diagonal matrix. 

Finally, we obtain the following action sequence update law:
\begin{align}
    \vu_t^{i+1} & = \vu_t^{i} + \sum_{k=0}^{K-1} w(\mathcal{E}^k) \bm{\epsilon}_t^{k} , \\
    \va_t^{i+1} & = \va_t^{i} + \vu_t^{i+1}{\Delta{t}} ,
\end{align}
and the trajectory cost (\ref{Equation:traj_cost}) now takes the form:
\begin{equation}
C(V^k, A) = S(V^k) + \Omega(A + V^k{\Delta{t}})+ \lambda\sum_{t=0}^{T-1} \vu_t^\rT \bm{\Sigma}^{-1} \bm{\epsilon}_t^k.
\end{equation}
Afterwards, we apply clamping function $\vg_{\va}$ to the action commands to impose box constraints on the vehicle's physical limits. The overall algorithm of SMPPI is shown in Alg.~\ref{Algorithm:smppi}.

\begin{algorithm}[tb]
\SetKwInOut{Input}{Given}
\Input{$\vF$: Dynamics model\;
       $K, T$: Number of samples, timesteps\;
       $(\vu_0, \vu_1, ... \, \vu_{T-1})$: Initial control sequence\;
       $(\va_0, \va_1, ... \, \va_{T-1})$: Initial action sequence\;
       $\bm{\Sigma}, \lambda, \phi, c, \Omega$: Control parameters and cost functions\;}     
\For{$i \leftarrow 0$ \KwTo \text{maximum iterations}}{
$\vx_0 \leftarrow \text{SubscribeState()}$\;
\For{$k \leftarrow 0$ \KwTo $K-1$}{
  $\vx \leftarrow \vx_0$\;
  Sample $\mathcal{E}^k = \left( \bm{\epsilon}_0^k \dots {\bm{\epsilon}_{T-1}^k} \right), ~\bm{\epsilon}_t^k \in \mathcal{N}(0, \bm{\Sigma})$\;
  \For{$t \leftarrow 0$ \KwTo $T-1$}{
    $\vv_{t}^{k} = \vu_{t} + \bm{\epsilon}_{t}^k$\;
    $\va_{t}^{k} = \va_{t} + \vv_t^{k} \Delta{t}$\;
    $\vx \leftarrow \vF(\vx, \, \va_{t}^{k})$\;
  	$C_k \pluseq c(\vx) + \lambda \vu_{t}^\rT \bm{\Sigma}^{-1} \bm{\epsilon}_t^k$\;
  }
  $C_k \pluseq \phi(\vx) + \Omega(\va_0^{k}, \va_1^{k}, \, \dots \, \va_{T-1}^{k})$\;
}

$w_k \leftarrow \text{ComputeWeights}(C_0, C_1, \dots C_{k-1})$\;
\For{$t \leftarrow 0$ \KwTo $T-1$}{
$U^{i+1} \leftarrow U^{i} + \left(\sum_{k=0}^{K-1} w_k^{\rT} \, \mathcal{E}^k\right)$\;
$A^{i+1} \leftarrow A^{i} + U^{i+1}{\Delta{t}}$\;
}

$\text{SendToController}(\va_0)$\;

\For{$t \leftarrow 1$ \KwTo $T-1$}{ 
	$\vu_{t-1}, \, \va_{t-1} \leftarrow \vu_t, \, \va_t$\;  
}
$\vu_{T-1}, \, \va_{T-1} \leftarrow \text{Initialize}(\vu_{T-1}, \va_{T-1})$\;
}
\caption{SMPPI \label{Algorithm:smppi}} 
\end{algorithm} 

\section{Experiments on an Inverted Pendulum}
\begin{figure}[t]
\centering
\includegraphics[width=0.99\linewidth]{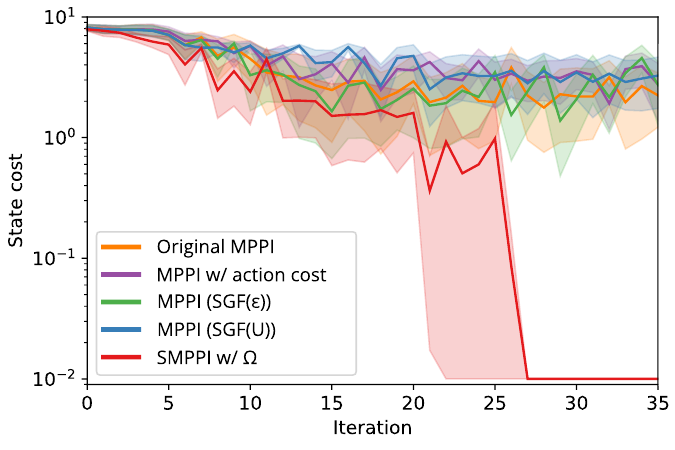}
\caption{Average state costs during optimization procedures with different control methods. Each iteration consists of $20$ time steps. Note that we apply logarithmic scaling to the y-axis and that the costs are clamped to a minimum of $0.01$. }
\label{fig:pendulum}
\end{figure}

We tested MPPI and SMPPI in a simulated inverted pendulum swing-up task. One of the important characteristics of the pendulum swing-up task is that the available torque is insufficient to push the pendulum to reach an upright position in a single rotation. The controller should swing the pendulum multiple times to gather energy, and during this process, the optimal control sequence changes rapidly. Consequently, it is a simple yet effective benchmark for evaluating our method. To exploit the fact that our idea is generalized to allow non-affine dynamics, we used a neural network to learn the dynamics model in this experiment.

We designed the dynamics model with two fully-connected hidden layers, each with 32 neurons. The state-action dataset was repeatedly collected during the control procedure and the model was trained with the dataset every 50 time steps. No bootstrap dataset was used because the dimensionality of the inverted pendulum system is sufficiently low. The running state cost function $c(\vx)$ in (\ref{Equation:state_cost}) is formulated as follows:
\begin{equation} 
    c(\vx \equiv [\theta, \dot{\theta}]^\rT) = \theta^2 + 0.1{\dot{\theta}^2} .
\label{Equation:pendulum_cost}
\end{equation}

We compared five different methods: \rnum{1}) the original MPPI without smoothing (abbreviated as ``MPPI baseline"), \rnum{2}) applying an additional action cost for smoothing on MPPI, \rnum{3}) applying the SGF on the noise sequence (abbreviated as ``MPPI ($\operatorname{SGF}(\bm{\epsilon})$)"), \rnum{4}) applying the SGF on the control sequence (abbreviated as ``MPPI ($\operatorname{SGF}(U)$)"), and \rnum{5}) SMPPI. In this task, control parameter $\lambda$ was set to 10. For SMPPI, the action sequence cost parameter $\bm{\omega}$ was set to 1. They were tested with seven different initial angular velocities ($\dot{\theta} \in \{-3, -2, \dots 3\}$) starting from the initial position vertically downward. We fixed the random seed for a fair comparison. The state costs (\ref{Equation:pendulum_cost}) during online optimization are shown in Fig.~\ref{fig:pendulum}.

The results show that only SMPPI successfully stabilized the pendulum in the upright position in all cases. As seen in our video, the other methods showed similar results. They continually failed to push the pendulum to the upright position because they did not apply enough torque during the swinging motions. In the case of MPPI baseline \rnum{1}), it was vulnerable to chattering and could not converge into an optimal control sequence when the optimal distribution changed rapidly. Imposing an additional action cost \rnum{2}), which was employed to alleviate chattering, also failed because this strategy violates the information theoretic interpretation of MPPI while using non-affine dynamics. Methods using external smoothing algorithms \rnum{3})$-$\rnum{4}) failed due to the negative effects of external smoothing. We discuss these issues further in Section~\ref{subsection:result}.

\section{Experiments on a Vehicle Simulator \label{section:carmaker}}
\subsection{Experimental Setup}

Our research goal is to perform high-speed driving successfully under challenging conditions, including those with unknown friction surfaces and sharp corners. The performance of the model-based control policy is highly dependent on the accuracy of the model \cite{nagabandi2018neural, sekar2020planning}. The vehicle model should be trained with all possible maneuvers to minimize any model bias. Because the control sequence update is processed by evaluating the perturbed control trajectories with sampling, situations in which control values change significantly over time that make the vehicle unstable should be included in the training dataset. Obtaining such data from an actual vehicle would be dangerous. Therefore, we used CarMaker to collect driving data and to evaluate the control performances. This is a widely used, high-fidelity vehicle simulator that precisely solves nonlinear dynamics in real time.

\begin{figure}[t]
\centering
\includegraphics[width=0.99\linewidth]{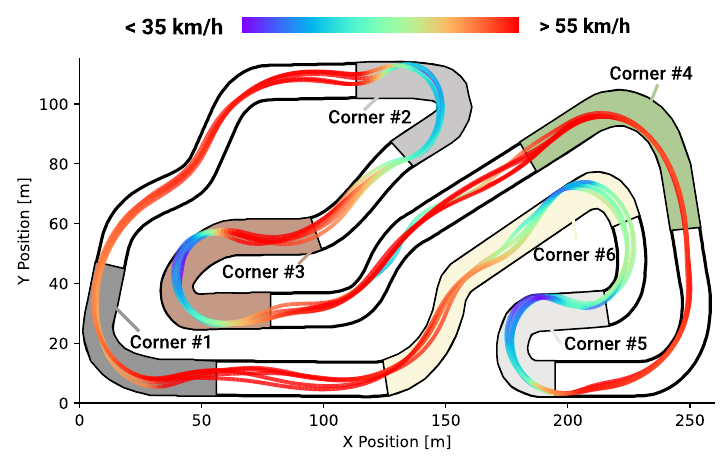}
\caption{Trajectory of our proposed method with a reference speed of $60$ km/h.}
\label{fig:full_trajectory}
\end{figure}

We built a race track modeled after an actual kart circuit known as ``KART 2000" in Kirchlengern, Germany (see Fig.~\ref{fig:full_trajectory}). The length of the track was $1016$ m and it had two moderate curves and four sharp curves. Note that when the vehicle reaches the entry of sharp corners, the controller should reduce the vehicle's speed and apply a large steering angle quickly. Otherwise, the vehicle will understeer and collide with the track boundary.

In some prior research on autonomous driving on race tracks, electric vehicles without transmissions were used \cite{williams2018information, kabzan2020amz, vazquez2020optimization}. Unlike these vehicles, the majority of current vehicles use an automatic transmission to shift gears. The dynamic characteristics of the vehicle will definitely change as a gear shifts. Therefore, throttle control is not feasible for use in common vehicles. Here, we simply overcome this issue by taking desired speed ($v_{des}$) as the feedforward command of MPPI. A low-level feedback controller then manages the throttle and brake via a Proportional-Integral (PI) loop. A Volvo XC90 was used as the control vehicle.

\subsection{Training the Neural Network Vehicle Model  \label{subsection:training}}

There are two main characteristics that make the vehicle dynamics challenging to model: First, vehicle motions under unknown friction are difficult to predict. In particular, the lateral tire forces will slip into the nonlinear region on low friction corners \cite{spielberg2019neural}. Second, the automatic gear shifting mechanism is highly complex to model explicitly. We propose to model such complex and nonlinear dynamics using a neural network with state-action history, which enables the network to capture time-varying behavior \cite{punjani2015deep}. The overall structure of our vehicle model, which is a feedforward fully-connected neural network, is shown in Table~\ref{table:model}. The terms $\left|\vx\right|$ and $\left|\va\right|$ denote correspondingly the size of the state and the action space, and $h$ represents the number of state-action pairs in the previous history. A Rectified Linear Unit (ReLU) is used for the activation of hidden layers. The Mean Squared Error (MSE) loss was used as the loss function and the Adam optimization was used for mini-batch gradient descent.

\begin{table}[ht]
\renewcommand\arraystretch{1.1}
\captionsetup{justification=centering}
\caption{Overall architecture of the vehicle model}
\label{table:model}
\begin{center}
\begin{tabular}{c|cc}
\, \, &  Size & Activation   \\ \hline
\,Input\,      & $(\left|\vx\right| + \left|\va\right|) \times h$         &  -      \\
\,Hidden Layer 1\,     & $2 \times$Input Size        &  ReLU      \\
\,Hidden Layer 2\,  & $4 \times$Input Size      &  ReLU     \\
\,Hidden Layer 3\,      & $6 \times$Input Size       & ReLU       \\
\,Hidden Layer 4\,      & $2 \times$Input Size    &  Linear    \\
\,Output\,      & $\left|\vx\right|$   &  -    \\
\end{tabular}
\end{center}
\vspace*{-0.15in}
\end{table}

We collected a human-controlled driving dataset in a manner similar to that utilized in our prior work on the training of a dynamics model \cite{bae2021curriculum}. The vehicle state is defined as $\vx = [v_x, v_y, r]^\rT$, where $v_x$ and $v_y$ are the longitudinal and lateral velocities, and $r$ is the yaw rate. The action command is defined as $\va = [\delta, v_{des}]^\rT$, where $\delta$ is the steering angle command. We carefully selected three distinct maneuvers: 
\begin{enumerate}
    \item Zig-zag driving at low speeds ($20-25$ km/h) on the race track.
    
    \item High-speed driving on the race track while trying to maintain a speed of $40$ km/h as much as possible.
    
    \item Sliding maneuvers during combinations of acceleration and deceleration with small, medium, and large steering angles. Subsequently, we controlled the vehicle with rapidly changing commands for random movements, representing the sampled noisy trajectories of MPPI. These maneuvers were performed on flat ground in both the left and right directions.
\end{enumerate}
Each maneuver was done with multiple road friction coefficients $\mu \in \{0.4, 0.5, \dots 1.0\}$. The maneuvers on the race track were done in both clockwise and counter-clockwise directions. Each one was logged for two minutes. We collected a total of 35 maneuvers, which amounted to 70 minutes of driving. The dataset was split into two portions: 70$\%$ for training and 30$\%$ for testing. To evaluate the generalization performance of the trained model, the validation dataset was collected on the same race track while the friction coefficients were modified. From \textit{Corner \#1} to \textit{Corner \#6} (depicted in Fig.~\ref{fig:full_trajectory}), the friction coefficients were assigned with values that were not included in the training dataset: $[0.95, 0.85, 0.75, 0.65, 0.55, 0.45]$, respectively. The other regions were assigned values of $\mu = 0.8$.

\begin{figure}[t]
\centering
\includegraphics[width=0.99\linewidth]{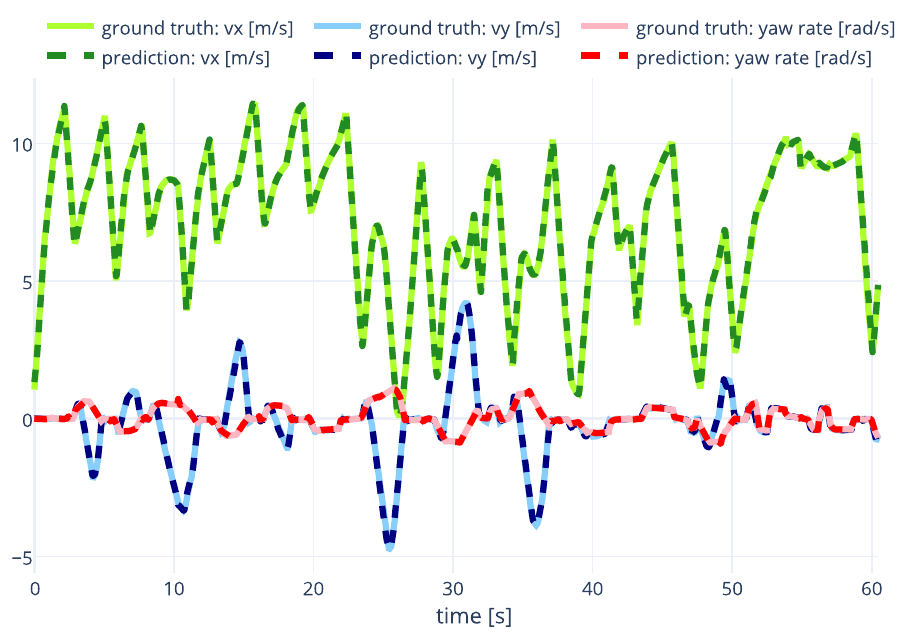}
\caption{Estimation results on the validation dataset.}
\label{fig:nn}
\end{figure}

The test and validation errors after training are shown in Table~\ref{table:errors}. The results show that our trained model can precisely represent the vehicle system dynamics regardless of the road friction. The estimation results on the validation data are visualized in Fig.~\ref{fig:nn}. The Root Mean Square Error (RMSE) is denoted as $\mathbf{E}_{RMS}$ and the max error is denoted as $\mathbf{E}_{max}$.

\begin{table}[ht]
\renewcommand\arraystretch{1.2}
\captionsetup{justification=centering}
\caption{Training results of the neural network vehicle model.}
\label{table:errors}
\begin{center}
\begin{tabularx}{1.0\columnwidth}{c|cc|cc|cc|}
& \multicolumn{2}{c|}{$v_x$ [m/s]} & \multicolumn{2}{c|}{$v_y$ [m/s]} & \multicolumn{2}{c|}{$r$ [rad/s]} \\ \cline{2-7}
& $\mathbf{E}_{RMS}$ & $\mathbf{E}_{max}$ & $\mathbf{E}_{RMS}$ & $\mathbf{E}_{max}$ & $\mathbf{E}_{RMS}$ & $\mathbf{E}_{max}$  \\ \cline{1-7}
\textbf{Test} & 0.0311 & 0.5530 & 0.0216 & 0.4723 & 0.0175 & 0.2506 \\
\textbf{Val.} & 0.0252 & 0.3744 & 0.0136 & 0.1500 & 0.0123 & 0.1400 \\
\end{tabularx}
\end{center}
\vspace*{-0.15in}
\end{table}

\subsection{Cost Function and Other Parameters}
Following prior work \cite{williams2018information}, we designed a state-dependent cost function $c(\vx)$, which consisted of three components:
\begin{equation} \label{eq:cost}
    c(\vx) = \alpha_1{\text{Track}(\vx)} + \alpha_2{\text{Speed}(\vx)} + \alpha_3{\text{Slip}(\vx)}.
\end{equation}
The track cost indicates whether the vehicle is inside the track or outside of the given boundary. The states of the sampled trajectories are transformed into global positions $(p_x, p_y)$ and they lookup the values of a two-dimensional cost map $\textbf{M}$, which represents the area of the outer boundary as 1 and the area of the inner boundary as 0. Our sampling-based controller allows this impulse-like cost function:
\begin{equation}
    \text{Track}(\vx) = (0.9)^t\,{10000\textbf{M}(p_x,p_y)} .
\end{equation}
The speed cost is the quadratic cost to achieve the reference vehicle speed:
\begin{equation}
    \text{Speed}(\vx) = (v_x - v_{ref})^2 .
\end{equation}
The slip cost penalizes the sideslip angle to plan a stable future trajectory. Additionally, it imposes a hard cost to reject samples which are estimated to have a larger sideslip angle than $0.2$ rad (approximately 11.46$^\circ$) in the future:
\begin{align}
    \text{Slip}(\vx) &= \sigma^2 + 10000I  \left(\{\left | \sigma \right | > 0.2\}\right) \\
    \sigma &= -\text{arctan} \left ( \frac{v_y}{\lVert v_x \rVert} \right ) , \nonumber
\end{align}
where $I$ is an indicator function.

We conducted a grid search for each controller to identify the best control parameters exhibiting the most successful performance. The sampling variance was adjusted to cover most of the control range of each controller. MPPI used $\bm{\Sigma} = \text{Diag}(4.0, 3.0)$ for the action noise and SMPPI used $\bm{\Sigma} = \text{Diag}(0.7, 0.4)$ for the derivative action noise. The feedforward control frequency was $10$ Hz and the controllers used a time horizon of $4$ s. The control parameters $K$ and $\lambda$ were correspondingly set to 10000 and 15.0 during the experiments. For SMPPI, it used $\bm{\omega} = \text{Diag}(0.8, 0.8)$.

\subsection{Experimental Results \label{subsection:result}}

We conducted an ablation study for our framework. The track used in these experiments is identical to that used to collect the validation data. We measured the average lap time taken by the vehicle to traverse each corner during five laps around the track, where the reference speed was set to $40$ km/h. If the vehicle went outside a boundary, it was placed at the start point and had to start a new lap. The same pre-trained vehicle model was used and there was no extra training during experiments. The results are shown in Table~\ref{table:time_took}. 

\begin{table}[ht]
\renewcommand\arraystretch{1.2}
\captionsetup{justification=centering}
\caption{Average lap times (in [s]) on six corners with different control methods. For SMPPI with $\Omega$, the minimum speed (in [km/h]) and the maximum slip angle (in [$^{\circ}$]) while passing through each corner are also analyzed below.}
\label{table:time_took}
\begin{center}
\begin{tabular}{c|cccccc}
Lap time [s]      & \#1 & \#2 & \#3 & \#4 & \#5 & \#6  \\ \hline
Original MPPI      & 11.08 & 9.60 & N/A & N/A & N/A & N/A     \\
MPPI ($\operatorname{SGF}(\bm{\epsilon})$) & 11.75 & 9.90 & N/A & N/A & N/A & N/A    \\
MPPI ($\operatorname{SGF}(U)$)      & 7.82 & N/A & N/A & N/A & N/A & N/A    \\
SMPPI w/o $\Omega$    & 8.39 & 7.71 & 11.64 & 8.77 & 9.86 & 13.11     \\
\rowcolor{gray!50} SMPPI w/ $\Omega$      & 7.36 & 7.66 & 10.83 & 8.47 & 9.39 & 12.86    \\  \hline \hline
\rowcolor{gray!50} Min. Speed [km/h]      & 24.71 & 20.37 & 23.30 & 36.51 & 28.24 & 29.77    \\ 
\rowcolor{gray!50} Max. Slip [$^{\circ}$]      & 5.34 & 11.09 & 9.64 & 2.91 & 5.32 & 4.94    \\ 
\end{tabular}
\end{center}
\vspace*{-0.1in}
\end{table}

\begin{figure}[t]
\centering
\includegraphics[width=0.99\linewidth]{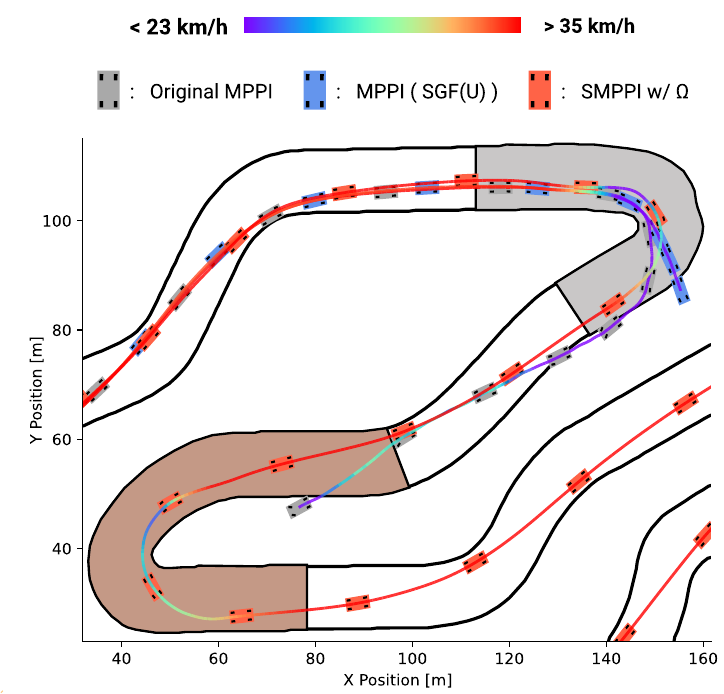}
\caption{Visualization of trajectories of the compared controllers. The friction coefficient of \textit{corner \#2} (in gray) is 0.85 and that of \textit{corner \#3} (in brown) is 0.75.}
\label{fig:trajectory_compare}
\vspace*{-0.1in}
\end{figure}

While all methods showed promising results on straight roads and on \textit{Corner \#1}, MPPI with different control update laws (\ref{Equation:no_filter}), (\ref{Equation:filter_on_noise}), (\ref{Equation:filter_on_u}) failed to complete \textit{Corner \#2} and \textit{Corner \#3}, which are the sharpest corners of the track. The vehicles collided with the boundary or lost control after entering these turns. In contrast, both of the proposed methods successfully controlled the vehicle to traverse all of the laps at high speeds. One notable finding is that SMPPI with the additional action cost $\Omega$ showed faster lap times in all corners than SMPPI without an action cost. This suggests that reducing chattering along the ``t-axis" can improve the control performance on the MPPI structure. Clearly, reducing the unnecessary noise in the steering and throttle commands is strongly related to increasing the speed of the vehicle. The trajectories taken by the three controllers are shown in Fig.~\ref{fig:trajectory_compare}. 

\begin{figure*}[t]
\centering
\includegraphics[width=0.99\textwidth]{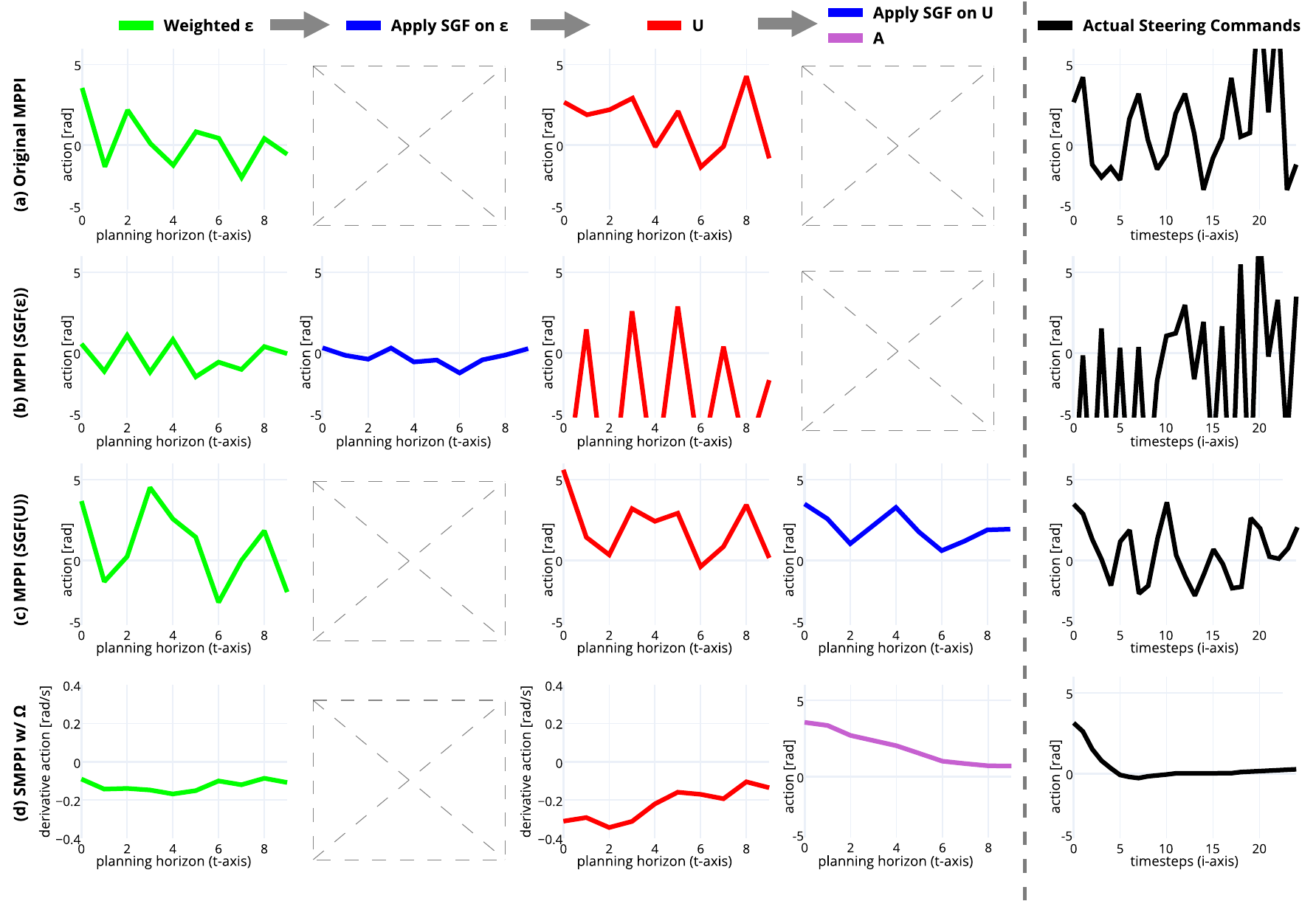}
\caption{The first 10 steps of the weighted sum of noise sequences (in green) and action sequences (in red) from different controllers right after entering \textit{Corner \#2} ($t=i=0$ at this instant). For clarity, only the steering angle commands are visualized. The results after applying the SGF filter are shown in blue. For SMPPI, the augmented action sequence is shown in purple. The actual steering commands applied to the vehicle for 25 steps ($2.5$ s) are shown on the right-hand side (in black).}
\label{fig:chattering}
\end{figure*}

Finally, we examined in more detail the actual optimization procedure of each method to verify our idea. The sequences computed by the four controllers right after taking \textit{Corner \#2} are visualized in Fig.~\ref{fig:chattering}. While passing through this sharp corner, the neural network was able to estimate that the vehicle would slide off the track if it did not slow down due to the friction limits. Therefore, the controller was required to apply the brakes and a large positive steering angle to make an appropriate right turn. During this procedure, the gap between the optimal and the current control sequence increased in an instant. The control sequence in the MPPI baseline attempted to respond to the rapidly changing optimal distribution. However, chattering occurred (see Fig.~\ref{fig:chattering}a) and caused the vehicle to lose its stability (see Fig.~\ref{fig:trajectory_compare}). This is because perturbed samples incorporating high-frequency noise survived after importance sampling, to minimize the impulse-like state cost despite the chattering. In the case of ``MPPI ($\operatorname{SGF}(\bm{\epsilon})$)", the external smoothing filter distorted the phase of the weighted noise. The noise sequence then escalated the chattering into the control sequence and finally led to divergence (see Fig.~\ref{fig:chattering}b). On the other hand, smoothing was suitably applied in ``MPPI ($\operatorname{SGF}(U)$)", but this rather hindered the controller from quickly generating the correct command (see Fig.~\ref{fig:chattering}c). This behavior is also well illustrated in Fig.~\ref{fig:trajectory_compare}. In contrast, SMPPI successfully responded to the environment and generated a smooth action sequence without the need for an additional smoothing filter. The derivative action sequence still possessed inevitable chattering, but it was integrated to become a smoother action sequence. The actual commands applied to the vehicle were also shown to be stable (along the ``i-axis"), an achievement realized due to the strong benefits of SMPPI.

\section{Conclusion}
We presented the Smooth Model Predictive Path Integral algorithm, which is designed to generate smooth control commands within the MPPI algorithm. Our method can attenuate chattering, which is a natural problem of sampling-based algorithms, without the need to design external smoothing filters. In particular, our method outperforms the MPPI baseline in cases in which the environment changes rapidly. In addition, our proposed method is not confined to employing action smoothing cost, as the control domain is shifted to the derivative action distribution. We showed that SMPPI can improve the general performance when it is applied to autonomous driving tasks. SMPPI was demonstrated to be capable of controlling an autonomous vehicle with agility on sharp and slippery corners. Furthermore, this chattering-free controller is also beneficial in reducing damage to the actuators.

\section*{APPENDIX}

We also demonstrated our method on a more aggressive driving task. The reference speed was set to $60$ km/h. We collected an extra training dataset of high-speed maneuvers on the race track, following the strategy described in Section~\ref{section:carmaker}. Then, we trained the vehicle model with the augmented dataset. Finally, the SMPPI controller was deployed on the validation race track. From \textit{Corner \#1} to \textit{Corner \#6} (depicted in Fig.~\ref{fig:full_trajectory}), the friction coefficients were assigned the following corresponding values: $[1.0, 0.95, 0.9, 0.85, 0.8, 0.75]$. The other regions were set such that $\mu = 0.9$. All five laps were completed at high speeds on challenging roads with varying amounts of friction. The trajectory taken by the vehicle is visualized in Fig.~\ref{fig:full_trajectory}. We were able to observe the controller taking ``out-in-out" trajectories on sharp corners because this is the best way to preserve high speeds and prevent large slip angles according to the predictions of the neural network model. We encourage readers to watch our supplementary video, where all of our experiments are shown in detail.

\bibliographystyle{IEEEtran}
\typeout{}
\bibliography{mybib.bib}

\begin{thebibliography}{10}
\providecommand{\url}[1]{#1}
\csname url@samestyle\endcsname
\providecommand{\newblock}{\relax}
\providecommand{\bibinfo}[2]{#2}
\providecommand{\BIBentrySTDinterwordspacing}{\spaceskip=0pt\relax}
\providecommand{\BIBentryALTinterwordstretchfactor}{4}
\providecommand{\BIBentryALTinterwordspacing}{\spaceskip=\fontdimen2\font plus
\BIBentryALTinterwordstretchfactor\fontdimen3\font minus
  \fontdimen4\font\relax}
\providecommand{\BIBforeignlanguage}[2]{{%
\expandafter\ifx\csname l@#1\endcsname\relax
\typeout{** WARNING: IEEEtran.bst: No hyphenation pattern has been}%
\typeout{** loaded for the language `#1'. Using the pattern for}%
\typeout{** the default language instead.}%
\else
\language=\csname l@#1\endcsname
\fi
#2}}
\providecommand{\BIBdecl}{\relax}
\BIBdecl

\bibitem{li2004iterative}
W.~Li and E.~Todorov, ``Iterative linear quadratic regulator design for
  nonlinear biological movement systems.'' in \emph{ICINCO}.\hskip 1em plus
  0.5em minus 0.4em\relax Citeseer, 2004, pp. 222--229.

\bibitem{tassa2014control}
Y.~Tassa, N.~Mansard, and E.~Todorov, ``Control-limited differential dynamic
  programming,'' in \emph{IEEE International Conference on Robotics and
  Automation (ICRA)}.\hskip 1em plus 0.5em minus 0.4em\relax IEEE, 2014, pp.
  1168--1175.

\bibitem{rosolia2020learning}
U.~Rosolia and F.~Borrelli, ``Learning how to autonomously race a car: A
  predictive control approach,'' \emph{IEEE Transactions on Control Systems
  Technology}, vol.~28, no.~6, pp. 2713--2719, 2020.

\bibitem{kabzan2020amz}
J.~Kabzan, M.~I. Valls, V.~J. Reijgwart, H.~F. Hendrikx, C.~Ehmke, M.~Prajapat,
  A.~B{\"u}hler, N.~Gosala, M.~Gupta, R.~Sivanesan \emph{et~al.}, ``Amz
  driverless: The full autonomous racing system,'' \emph{Journal of Field
  Robotics}, vol.~37, no.~7, pp. 1267--1294, 2020.

\bibitem{williams2017information}
G.~Williams, N.~Wagener, B.~Goldfain, P.~Drews, J.~M. Rehg, B.~Boots, and E.~A.
  Theodorou, ``Information theoretic mpc for model-based reinforcement
  learning,'' in \emph{IEEE International Conference on Robotics and Automation
  (ICRA)}.\hskip 1em plus 0.5em minus 0.4em\relax IEEE, 2017, pp. 1714--1721.

\bibitem{williams2018information}
G.~Williams, P.~Drews, B.~Goldfain, J.~M. Rehg, and E.~A. Theodorou,
  ``Information-theoretic model predictive control: Theory and applications to
  autonomous driving,'' \emph{IEEE Transactions on Robotics}, vol.~34, no.~6,
  pp. 1603--1622, 2018.

\bibitem{yinTrajectoryDistributionControl2022}
J.~Yin, Z.~Zhang, E.~Theodorou, and P.~Tsiotras, ``Trajectory {{Distribution
  Control}} for {{Model Predictive Path Integral Control}} using {{Covariance
  Steering}},'' in \emph{IEEE International {{Conference}} on {{Robotics}} and
  {{Automation}} ({{ICRA}})}, 2022.

\bibitem{vazquez2020optimization}
J.~L. V{\'a}zquez, M.~Br{\"u}hlmeier, A.~Liniger, A.~Rupenyan, and J.~Lygeros,
  ``Optimization-based hierarchical motion planning for autonomous racing,'' in
  \emph{IEEE/RSJ International Conference on Intelligent Robots and Systems
  (IROS)}.\hskip 1em plus 0.5em minus 0.4em\relax IEEE, 2020, pp. 2397--2403.

\bibitem{chisari2021learning}
E.~Chisari, A.~Liniger, A.~Rupenyan, L.~Van~Gool, and J.~Lygeros, ``Learning
  from simulation, racing in reality,'' in \emph{IEEE International Conference
  on Robotics and Automation (ICRA)}.\hskip 1em plus 0.5em minus 0.4em\relax
  IEEE, 2021, pp. 8046--8052.

\bibitem{kobilarov2012cross}
M.~Kobilarov, ``Cross-entropy motion planning,'' \emph{The International
  Journal of Robotics Research}, vol.~31, no.~7, pp. 855--871, 2012.

\bibitem{theodorou2012relative}
E.~A. Theodorou and E.~Todorov, ``Relative entropy and free energy dualities:
  Connections to path integral and kl control,'' in \emph{IEEE Annual
  Conference on Decision and Control (CDC)}.\hskip 1em plus 0.5em minus
  0.4em\relax IEEE, 2012, pp. 1466--1473.

\bibitem{sarkka2008unscented}
S.~S{\"a}rkk{\"a}, ``Unscented rauch--tung--striebel smoother,'' \emph{IEEE
  Transactions on Automatic Control}, vol.~53, no.~3, pp. 845--849, 2008.

\bibitem{ruiz2017particle}
H.-C. Ruiz and H.~J. Kappen, ``Particle smoothing for hidden diffusion
  processes: Adaptive path integral smoother,'' \emph{IEEE Transactions on
  Signal Processing}, vol.~65, no.~12, pp. 3191--3203, 2017.

\bibitem{savitzky1964smoothing}
A.~Savitzky and M.~J. Golay, ``Smoothing and differentiation of data by
  simplified least squares procedures.'' \emph{Analytical Chemistry}, vol.~36,
  no.~8, pp. 1627--1639, 1964.

\bibitem{adhikari2020physiological}
K.~Adhikari, S.~Tatinati, K.~C. Veluvolu, and J.~A. Chambers, ``Physiological
  tremor filtering without phase distortion for robotic microsurgery,''
  \emph{IEEE Transactions on Automation Science and Engineering}, 2020.

\bibitem{wang2014fractional}
J.~Wang, Y.~Ye, X.~Pan, X.~Gao, and C.~Zhuang, ``Fractional zero-phase
  filtering based on the riemann--liouville integral,'' \emph{Signal
  Processing}, vol.~98, pp. 150--157, 2014.

\bibitem{nakatani2019swing}
S.~Nakatani and H.~Date, ``Swing up control of inverted pendulum on a cart with
  collision by monte carlo model predictive control,'' in \emph{Annual
  Conference of the Society of Instrument and Control Engineers of Japan
  (SICE)}.\hskip 1em plus 0.5em minus 0.4em\relax IEEE, 2019, pp. 1050--1055.

\bibitem{williams2017model}
G.~Williams, A.~Aldrich, and E.~A. Theodorou, ``Model predictive path integral
  control: From theory to parallel computation,'' \emph{Journal of Guidance,
  Control, and Dynamics}, vol.~40, no.~2, pp. 344--357, 2017.

\bibitem{nagabandi2018neural}
A.~Nagabandi, G.~Kahn, R.~S. Fearing, and S.~Levine, ``Neural network dynamics
  for model-based deep reinforcement learning with model-free fine-tuning,'' in
  \emph{IEEE International Conference on Robotics and Automation (ICRA)}.\hskip
  1em plus 0.5em minus 0.4em\relax IEEE, 2018, pp. 7559--7566.

\bibitem{sekar2020planning}
R.~Sekar, O.~Rybkin, K.~Daniilidis, P.~Abbeel, D.~Hafner, and D.~Pathak,
  ``Planning to explore via self-supervised world models,'' in
  \emph{International Conference on Machine Learning (ICML)}.\hskip 1em plus
  0.5em minus 0.4em\relax PMLR, 2020, pp. 8583--8592.

\bibitem{spielberg2019neural}
N.~A. Spielberg, M.~Brown, N.~R. Kapania, J.~C. Kegelman, and J.~C. Gerdes,
  ``Neural network vehicle models for high-performance automated driving,''
  \emph{Science Robotics}, vol.~4, no.~28, 2019.

\bibitem{punjani2015deep}
A.~Punjani and P.~Abbeel, ``Deep learning helicopter dynamics models,'' in
  \emph{IEEE International Conference on Robotics and Automation (ICRA)}.\hskip
  1em plus 0.5em minus 0.4em\relax IEEE, 2015, pp. 3223--3230.

\bibitem{bae2021curriculum}
J.~Bae, T.~Kim, W.~Lee, and I.~Shim, ``Curriculum learning for vehicle lateral
  stability estimations,'' \emph{IEEE Access}, vol.~9, pp. 89\,249--89\,262,
  2021.

\end{thebibliography}

\end{document}